\documentclass[10pt,twocolumn,letterpaper]{article}

\usepackage{cvpr}
\usepackage{amsmath}  
\usepackage{mathrsfs} 

\usepackage{pifont}
\newcommand{\red}[1]{{\color{red}#1}}
\newcommand{\blue}[1]{{\color{blue}#1}}

\definecolor{cvprblue}{rgb}{0.21,0.49,0.74}
\usepackage[pagebackref,breaklinks,colorlinks,allcolors=cvprblue]{hyperref}
\usepackage{multirow}

\title{PP-SSL: Priority-Perception Self-Supervised Learning \\ for Fine-Grained Visual Recognition}

\author{
   Shuaiheng Li\textsuperscript{1}~~~Qing Cai\textsuperscript{1}~~~Fan Zhang\textsuperscript{2}~~~Menghuan Zhang\textsuperscript{1}~~~Yangyang Shu\textsuperscript{4}\\
   Zhi Liu\textsuperscript{3}~~~Huafeng Li\textsuperscript{5}
   ~~~Lingqiao Liu\textsuperscript{4} \\
     \textsuperscript{1}College of  Computer Science and Technology, Ocean University of China,\\
     \textsuperscript{2}School of Automation, Northwestern Polytechnical University \\
     \textsuperscript{3}School of Information Science and Engineering, Shandong University \\
    \textsuperscript{4}School of Computer Science, The University of Adelaide \\
    \textsuperscript{5}Faculty of Information Engineering and Automation, Kunmimg University of Science and Technology\\
\texttt{\{lsh3567, zhangmenghuan\}@stu.ouc.edu.cn, cq@ouc.edu.cn, fanz6095@gmail.com}
}

\let\oldtwocolumn\twocolumn
\renewcommand\twocolumn[1][]{%
    \oldtwocolumn[{#1}{
    \begin{center}
           \includegraphics[width=\textwidth]{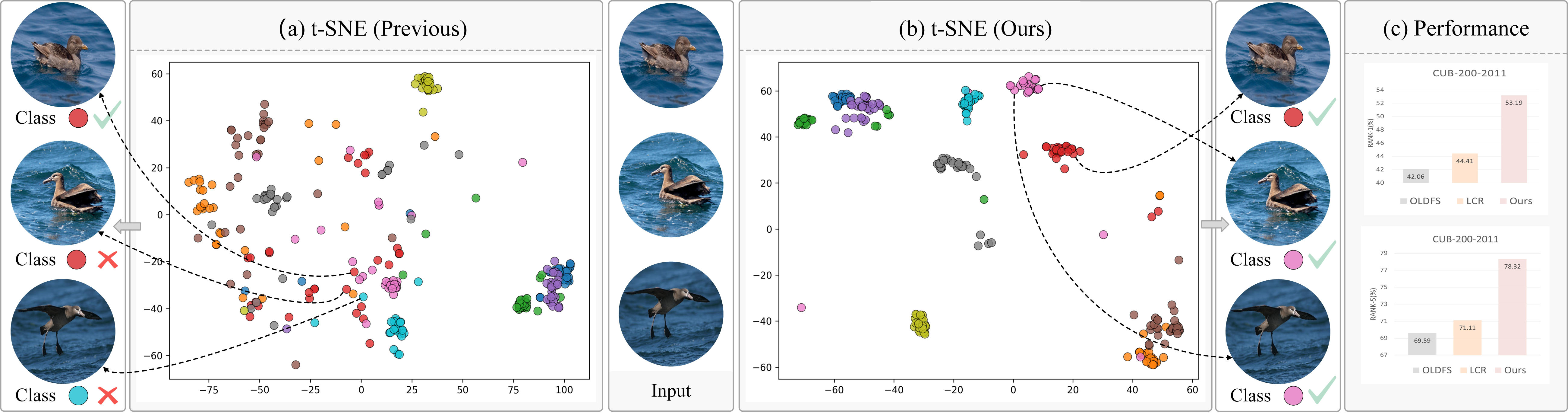}
           \captionof{figure}{{As shown in the input images, the top image differs from the bottom two, which belong to the same category but exhibit subtle inter-class differences and large intra-class variations. The top two images have similar backgrounds and poses, leading to potential misclassification as the same class. The bottom two images, despite being from the same class, have significant pose variations, causing misclassification. As shown in (a), previous methods struggle with poor category separation due to these factors. In contrast, (b) demonstrates that our method improves feature discriminability by using the Anti-Interference Strategy (AIS) to filter irrelevant features and the Image-Aided Distinction Module (IADM) to focus on fine-grained details, significantly enhancing category separation and recognition. (c) shows significant improvements of our method in rank-1 and rank-5 accuracy on the CUB-200-2011 dataset.}}
           \label{fig1}
        \end{center}
    }]
}

\begin{document}
\maketitle

\begin{abstract}
Self-supervised learning is emerging in fine-grained visual recognition with promising results. However, existing self-supervised learning methods are often susceptible to irrelevant patterns in self-supervised tasks and lack the capability to represent the subtle differences inherent in fine-grained visual recognition (FGVR), resulting in generally poorer performance. To address this, we propose a novel Priority-Perception Self-Supervised Learning framework, denoted as PP-SSL, which can effectively filter out irrelevant feature interference and extract more subtle discriminative features throughout the training process. Specifically, it composes of two main parts: the Anti-Interference Strategy (AIS) and the Image-Aided Distinction Module (IADM). In AIS, a fine-grained textual description corpus is established, and a knowledge distillation strategy is devised to guide the model in eliminating irrelevant features while enhancing the learning of more discriminative and high-quality features. IADM reveals that extracting GradCAM from the original image effectively reveals subtle differences between fine-grained categories. Compared to features extracted from intermediate or output layers, the original image retains more detail, allowing for a deeper exploration of the subtle distinctions among fine-grained classes. Extensive experimental results indicate that the PP-SSL significantly outperforms existing methods across various datasets, highlighting its effectiveness in fine-grained recognition tasks. Our code will be made publicly available upon publication.

\end{abstract}    
\section{Introduction}

Self-supervised learning (SSL) \cite{wei2022can,deng2022insclr1,bardes2021vicreg} have demonstrated impressive performance in various visual tasks like image classification \cite{jaiswal2020survey}, object detection \cite{wang2021dense}, semantic segmentation \cite{caron2021emerging} and image retrieval \cite{deng2022insclr1}, enabling models to capture general feature representations without labeled data. Recently, an increasing number of self-supervised methods have been proposed, which can be roughly categorized into two groups: clustering-based methods \cite{caron2018deep, guo2018deep, yang2016joint,chang2017deep} and contrastive learning-based methods \cite{chen2020simple, he2020mocov1, grill2020bootstrap,caron2020unsupervised}. 

Clustering-based methods learn the structure of data by grouping it into different clusters or groups. However, it can not effectively optimize inter-class distances through positive and negative sample pairs \cite{chen2020simple, he2020mocov1}. In contrast, contrastive learning-based methods demonstrate superior feature learning capabilities by learning data representations through comparisons between positive and negative samples. Owing to its notable performance, several researchers employ it in FGVR tasks~\cite{zbontar2021barlow,liu2021self,zhang2024self}, and have achieved impressive performance. Different from the research on large-scale general image datasets \cite{russakovsky2015imagenet,thomee2016yfcc100m,deng2009imagenet}, FGVR tasks require to differentiate subtle visual patterns, and primarily focuses on identifying subcategories within visual data, such as different bird species \cite{berg2014birdsnap,van2015building,wah2011caltech}, aircraft models \cite{maji2013fine}, and vehicle types~\cite{krause20133d}. Therefore, existing contrastive learning-based methods may suffer from ``granularity gap" (i.e., the disparity between coarse-grained and fine-grained features) \cite{cole2022does}. Moreover, recent studies show that existing methods are usually distracted by irrelevant features (i.e., the background noise)~\cite{kim2023coreset,shu2023learning,shu2022improving}, resulting in feature entanglement in FGVR tasks and suboptimal intra-class boundaries (see Fig.~\ref{fig1}). 

To address these challenges, we propose a novel priority perception self-supervised learning framework, which effectively solves the issues of irrelevant feature interference and mitigating granularity bias. Specifically, the proposed Anti-Interference Strategy (AIS) leverages the unique decoupled modality property of CLIP \cite{CLIP} by embedding fine-grained text representations. In the fine-grained text corpus, we define both relevant and irrelevant items to the current task, which are stored as shared embeddings. This process guides the image encoder to filter out interference from irrelevant features, allowing it to extract meaningful visual representations rather than relying solely on image-level features. By implementing this strategy, we eliminate interference from irrelevant features without depending on labeled data, thereby facilitating seamless integration into the self-supervised learning training process. Furthermore, we assert that the original image retains the most comprehensive details. Our findings indicate that leveraging information from the original image can assist the network in learning subtle distinctions between categories. Consequently, we designed the Image-Aided Distinction Module (IADM), which focuses on capturing crucial details to mitigate the impact of subtle inter-class differences and large intra-class variations, which generates GradCAM \cite{selvaraju2017grad} by taking gradients of the original image with respect to the contrastive learning loss \cite{hadsell2006dimensionality}, allowing us to identify important regions within the original image. This guides the network's attention to focus on these regions, facilitating the exploration of more nuanced discriminative representations. During the inference phase, we eliminate redundant modules to maintain a streamlined and lightweight process, relying solely on the image encoder for predictions and generating features for downstream tasks. Extensive experimental results demonstrate that our proposed method significantly enhances the performance of self-supervised learning in fine-grained recognition tasks. 

Our main contributions are summarized as follows:
\begin{itemize}
\item We propose a self-supervised learning framework tailored for fine-grained recognition, with experimental results demonstrating its effectiveness on benchmark datasets and significant performance improvements in both retrieval and classification tasks.

\item We propose an Anti-Interference Strategy (AIS) that leverages a fine-grained text corpus to mitigate the interference of irrelevant features, thereby facilitating the model's learning of high-quality visual representations that are crucial for the task.

\item We design the Image-Aided Distinction Module (IADM) to extract fine-grained cues from the original images. By leveraging this information, the network learns subtle category distinctions, mitigating the impact of inter-class differences and intra-class variations. This approach guides the network to focus on more discriminative regions, offering a novel perspective for fine-grained tasks.
\end{itemize}

\begin{figure*}[!t] 
    \centering
    \includegraphics[width=0.99\textwidth]{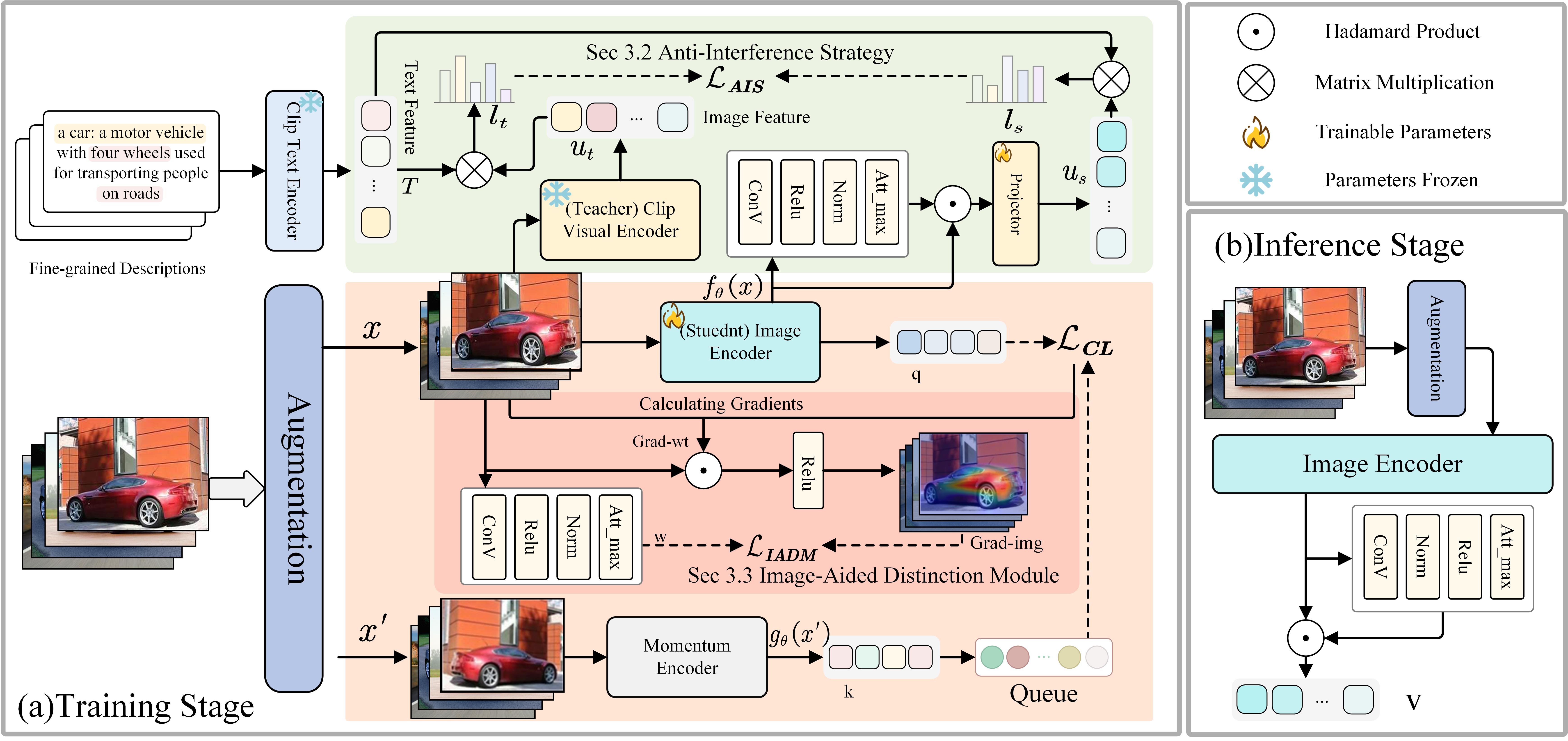}
    \caption{(a) Overview of our self-supervised framework: By incorporating AIS and IADM during the self-supervised training process, we effectively address the issue of irrelevant feature interference and extract the most detailed discriminative cues from the original images, thereby improving the performance of self-supervised learning in fine-grained recognition tasks. (b) During the inference phase, we remove redundant components, requiring only the output from the image encoder to be applied to downstream tasks, offering enhanced flexibility and convenience.}
    \label{Pipline}
\end{figure*}
\section{Related Works}
\textbf{Self-Supervised Learning (SSL)} has made significant progress in the field of computer vision by designing pretext tasks to learn useful feature representations from unlabeled data. Early methods, such as Jigsaw \cite{doersch2015unsupervised} and Jigsaw++ \cite{noroozi2016unsupervised}, learn feature representations by shuffling and restoring image patches, effectively improving image feature learning. In recent years, contrastive learning has become an important direction in self-supervised learning. The MoCo \cite{he2020mocov1} achieves efficient feature learning by building a dynamic dictionary and contrastive learning. The SimCLR \cite{chen2020simple} learns image features through data augmentation and a contrastive loss function. The BYOL \cite{grill2020bootstrap}, which conducts contrastive learning in a self-guided manner without the need for negative samples. Additionally, SwAV \cite{caron2020unsupervised} implements self-supervised learning by swapping cluster assignments across different views. MAE (Masked Autoencoders) \cite{he2022masked} learn feature representations effectively by masking parts of the input data and predicting the masked parts. 

\noindent \textbf{Self-Supervised Learning for Fine-Grained Visual Recognition.} Despite the impressive transferability and generalization demonstrated by SSL methods in many tasks, recent studies \cite{kim2023coreset,cole2022does} pointed out that it is hard to capture critical features for fine-grained visual recognition. To overcome this, researchers have proposed several improvements.
On the one hand, some methods focus on improving data augmentation techniques. For instance, DiLo \cite{zhao2021distilling} generates images with different backgrounds by combining images with new backgrounds, thereby enhancing the model's ability to localize foreground objects. ContrastiveCrop \cite{peng2022crafting} introduces an optimized cropping method to generate better views of the image. OLDFS \cite{wang2024learning} enhances the discriminative capability of the encoder by perturbing feature vectors to generate realistic synthetic images. 
On the other hand, Researchers aim to enhance the encoder's focus on salient regions by linking auxiliary neural networks to its convolutional layers. For example, CAST \cite{selvaraju2021casting} aligns Grad-CAM attention with key regions from saliency detectors to improve feature learning. CVSA \cite{di2021align} generates new views by cropping and swapping salient regions and employs cross-view saliency alignment loss to focus on foreground features. Nonetheless, they typically depend on pre-trained saliency detectors. LCR \cite{shu2023learning} and SAM \cite{shu2022improving} eliminate the dependence on pre-trained saliency detectors by guiding the network to match Grad-CAM outputs, with Grad-CAM serving as a benchmark for aligning the encoder's attention maps.

Despite significant advances in self-supervised learning for fine-grained visual recognition, several challenges remain. Irrelevant factors, such as background clutter, often obscure subtle feature differences, making it difficult to discern fine-grained distinctions. Additionally, small inter-class variations, coupled with large intra-class discrepancies, further complicate accurate recognition. These issues highlight the critical need for methods that can both minimize interference and effectively extract nuanced features. Addressing these challenges is essential for achieving more precise and robust fine-grained visual recognition.
\section{Method}
As illustrated in Fig.~\ref{Pipline}, we propose a Priority-Perception Self-Supervised Learning framework, which mainly consists of two key components: the Anti-Interference Strategy (AIS) and the Image-Aided Distinction Module (IADM).

\subsection{Preliminary}
Given an image $I$ from a batch of samples, two different data augmentation operations are applied to introduce perturbations, resulting in images $x$ and $x'$. These augmented images are then processed through the image encoder $f_\theta$ and momentum encoder $g_\theta$ to obtain feature embeddings $q$ and $k$. $q = f_\theta(x)$ and $k=g_\theta(x^{'})$. The $q$ and $k$, derived from the same image, serve as positive pairs. Conversely, embeddings $\{k_1, k_2, k_3, \ldots \}$, obtained from different views of other images, serve as negative pairs and are stored in a queue as a negative sample pool. Consequently, we can compute the contrastive learning loss \cite{chen2020improved} for the first stage:
\begin{equation}
\mathcal{L}_{\text{CL}}(q,k) = -\log \frac{\exp(q \cdot k / \tau)}{\sum_{i=1}^{K} \exp((1 \cdot k_i / \tau)},
\label{eq:contrastive_loss}
\end{equation}
where $\tau$ is the temperature parameter. $k$ is the number of negative samples in the queue.

\subsection{Anti-Interference Strategy (AIS)}

In our implementation, we regard the image encoder within the contrastive learning framework as the student model and the CLIP image encoder as the teacher model, with the CLIP text encoder serving as the bridge between the two. Given the nature of our task, our objective is to enable the network to distinguish irrelevant feature interference during the self-supervised learning process. 

To achieve this, as shown in Fig.~\ref{Pipline}, we have pre-designed a fine-grained textual corpus that includes attribute descriptions for several common categories, along with broader category descriptions relevant to the fine-grained datasets employed in this paper. It aims to enable the student image encoder to recognize these attributes, thereby filtering out irrelevant feature interference. We have designed eight fine-grained attribute descriptions, denoted as $t = {text_i}^N_{i=1}$, where $N=8$. These descriptions include examples such as ``an animal characterized by feathers, wings, and the ability to fly or perch." Among these, seven descriptions are unrelated to the current image, while one is relevant, encouraging the model to learn the ability to filter out irrelevant features and achieve high-quality feature extraction. We input the text corpus into the CLIP text encoder to obtain text embeddings $T \in \mathbb{R}^{N \times d}$, which are then $L2$ normalized. These text embeddings $T$ serve as shared feature representations between the student image encoder and the teacher CLIP image encoder. Based on this strategy, we only need to train the student image encoder. By inputting the images $x$ from the unlabeled training dataset $D_{u}$ into the pre-trained teacher CLIP image encoder, we obtain the normalized image embedding $u_t = {{f^t_I(x)}/{||f^t_I(x)||_2}} \in \mathbb{R}^d$.

After obtaining the visual embedding $f_\theta(x)$ from the student image encoder, we first extract the feature map using a $1 \times 1$ convolution kernel, followed by further processing to generate the visual features required for distillation:
\begin{equation}
z^{'} = max(norm(relu(\psi(f(x)))),
\label{eq:3}
\end{equation}
\begin{equation}
    u_s = Projector(f_\theta(x) \odot z^{'}),
\end{equation}
where $\psi(\cdot)$ denotes a 1 $\times$ 1 convolution kernel, $norm(\cdot)$ is defined as $\alpha_{i, j}^{\prime}=\frac{\alpha_{i, j}-\min (\alpha)}{1 \times 10^{-7} + \max (\alpha)}$, and $relu(\cdot)$ represents the ReLU activation function.  while $max(\cdot)$ represents the max-out operation, which selects the maximum value across each channel. The symbol $\odot$ indicates the Hadamard product. Finally, a learnable $projector(\cdot)$, is applied to ensure efficient and precise alignment with $u_t$ while maintaining low computational overhead, yielding the student image embedding $u_s$. The student image embedding ${{u_s}/{||u||_2}} \in \mathbb{R}^d$ is obtained by performing matrix multiplication with the text embedding $T \in \mathbb{R}^{N \times d}$ to obtain the logits $l_s = u_s  T^T \in \mathbb{R}^N$. Similarly, the image embedding $u_t \in \mathbb{R}^d$ from the CLIP image encoder is also multiplied by the text embedding $T \in \mathbb{R}^{N \times d}$ to generate the logits $l_t = u_t  T^T \in \mathbb{R}^N$. By optimizing our image encoder, we aim to produce image embeddings with semantic understanding capabilities on the unlabeled dataset $D_{u}$, thereby reducing the influence of irrelevant feature interference.

\begin{figure}[!t]
    \centering
    \includegraphics[width=\linewidth]{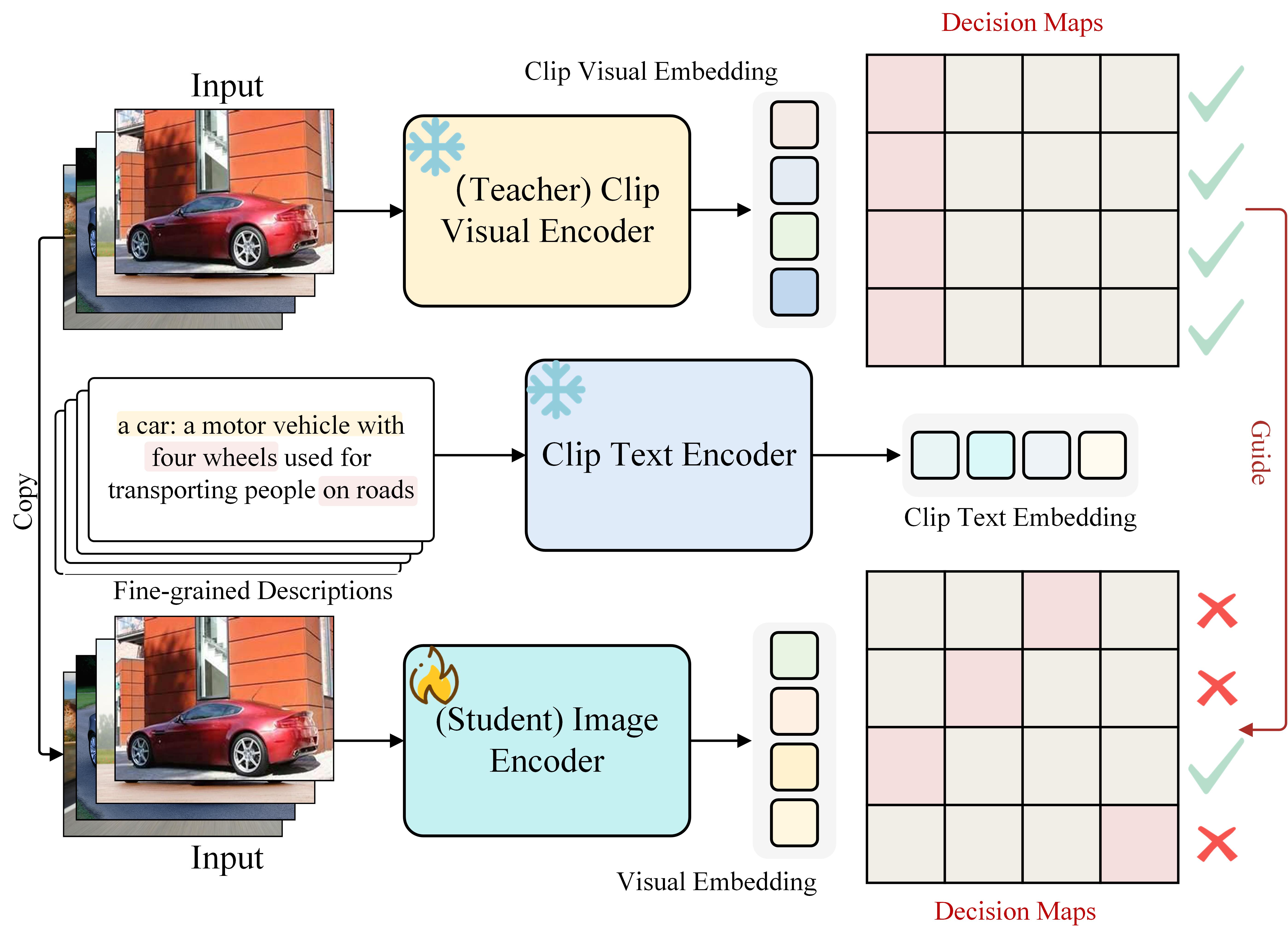}
    \caption{Our AIS utilizes the CLIP image encoder to guide the encoder in generating high-quality features with semantic category understanding. In the diagram, input images are all of the ``Cars" category, with one relevant attribute description and other unrelated descriptions in the text. This setup constrains the model to produce high-quality, semantically aware representations.}
    \label{AIS}
\end{figure}

The distillation process of AIS is illustrated in Fig.~\ref{AIS}. Knowledge distillation, first introduced by Hinton \cite{hinton2015distilling}, uses Kullback-Leibler (KL) divergence \cite{kullback1951information} to align outputs, optimizing the following objective:
\begin{equation}
    \mathbf{L}_{AIS}(l_t,l_s,\tau) = \tau^2KL(\sigma(l_t/\tau), \sigma(l_s/\tau)),
\end{equation}
where $l_t$ and $l_s$ denote the predictable logits of teacher model and student model. $\sigma(\cdot)$ denotes the softmax function, $\tau$ is the temperature parameter, which control the smoothness of the distributions.

\subsection{Image-Aided Distinction Module (IADM)}
Our IADM method is designed based on the GradCAM technique, which computes gradients with respect to the original image.

By substituting the cross-entropy loss in the standard GradCAM computation with the contrastive loss derived from Eq.~\ref{eq:contrastive_loss}, we can compute the GradCAM as follows:

 \begin{equation}
    Grad\text{-}wt = \left({\frac{\partial \mathcal{L}_{C L}(f_\theta(x),g_\theta(x^{'}))}{\partial x}}^{\top}\right),
    \label{eq:Grad_wt}
\end{equation}
 \begin{equation}
    Grad\text{-}Img = ReLU\left(Grad\text{-}{wt} \odot x \right).
    \label{eq:GradCAM}
\end{equation}

Eq.~\ref{eq:Grad_wt} computes the importance of each region in the image through gradients derived from the contrastive loss in the self-supervised learning framework, which is used to calculate the GradCAM weights. In Eq.~\ref{eq:GradCAM}, the GradCAM weights are multiplied with the original image to obtain the final GradCAM visualization. This serves as a pseudo-label for the regions of interest, guiding the network to focus on subtle details in the original image. Subsequently, we apply the same series of operations on the original image $x$ as in the AIS framework, as described below:
\begin{equation}
w = max(norm(relu(\psi(x)))),
\end{equation}
where remaining operations follow a similar procedure to those in AIS.
Our optimization objective is as follows:
\begin{equation}L_{\mathrm{IADM}}(Grad\text{-}Img\parallel w)=Grad\text{-}Img \cdot \log\frac{Grad\text{-}Img}{w},
\end{equation}
where symbol $\cdot$ denotes multiplication operation.

\subsection{Total Loss and Inference}
Overall, the loss function during training can be defined as:
\begin{equation}
   \mathcal{L}_{total} =\mathcal{L}_{CL} + \alpha \mathcal{L}_{AIS} + \beta \mathcal{L}_{IADM} ,
   \label{total+loss}
\end{equation}
where $\alpha=1.2$, and $\beta=0.01$ denote the hyperparameters that control the weight of the loss function.

During inference, the additional computations required during the training phase are no longer needed. We use the image encoder $f(\cdot)$ to generate the image embedding $f(x)$. 
The final features $f$ applied to the downstream task are obtained through the following operations:
\begin{equation}
v = normalize(AvgPool(z^{'} \odot f(x))),
\end{equation}
where $z^{'}$ denotes the result derived from Eq. \ref{eq:3}, with $AvgPool(\cdot)$ representing the average pooling operation and $normalize(\cdot)$ indicating the L2 normalization operation. As shown in Fig.~\ref{Pipline} (b), $v$ is used for downstream tasks.

\section{Experiments}

\begin{table*}[!t]
\centering
\renewcommand{\arraystretch}{1.2}  
\caption{All models use ResNet-50 as the network backbone, with the ResNet-50 architecture initialized using ImageNet-trained weights. We conduct a comparison with current state-of-the-art methods (i.e., LCR \cite{shu2023learning} and OLDFS \cite{wang2024learning}) on three benchmark datasets: CUB-200-2011, Stanford Cars, and FGVC Aircraft. For both the retrieval and classification tasks, the batch size is set to 128. The results for retrieval accuracy, rank-1, rank-5, and mAP (all in \%) are reported. For classification tasks, results are reported on 3 different label proportions: 100\%, 50\%, and 20\%. The best results are highlighted in \red{red}, and the second-best results are highlighted in \blue{blue}.}
\setlength{\tabcolsep}{4mm} 
\label{tab:1}
\resizebox{\textwidth}{!}{
\begin{tabular}{c|c|ccc|ccc}
\toprule
\multirow{2}{*}{DataSet} & \multirow{2}{*}{Method} & \multicolumn{3}{c|}{Retrieval} & \multicolumn{3}{c}{Classification} \\
\cmidrule(lr){3-5} \cmidrule(lr){6-8}
 &  & rank-1 & rank-5 & mAP & Top 1 / Top 5 (100) & Top 1 / Top 5 (50) & Top 1 / Top 5 (20) \\
\midrule
\multirow{3}{*}{CUB-200-2011} 
& LCR \cite{shu2023learning} & \blue{44.41} & \blue{71.11} & \blue{20.43} & 65.19 / 89.25 & 58.15 / 83.33 & 44.82 / 76.46 \\
& OLDFS \cite{wang2024learning} & 42.06 & 69.59 & 19.70 & \blue{66.17} / - & \blue{60.84} / - & \blue{49.69} / - \\
& PP-SSL (Ours) & \red{53.19} & \red{78.32} & \red{26.31} & \red{69.26 / 91.23} & \red{63.03/88.02} & \red{52.49 / 80.95} \\
\midrule
\multirow{3}{*}{Stanford Cars} 

& LCR \cite{shu2023learning} & \blue{36.46} & \blue{63.00} & 9.28 & 65.54 / 88.50 & \blue{54.77 / 81.84} & 36.46 / 65.86 \\
& OLDFS \cite{wang2024learning} & 35.81 & 61.94 & \blue{10.02} & \blue{65.60} / - & 54.36 / - & \blue{40.24} / - \\
& PP-SSL (Ours) & \red{41.25} & \red{68.25} & \red{11.37} & \red{67.73 / 90.15} & \red{57.73 / 84.40} & \red{40.99 / 70.22} \\
\midrule
\multirow{3}{*}{FGVC Aircraft} 
& LCR \cite{shu2023learning} & 32.97 & \blue{58.42} & 12.12 & 54.01 / 83.91 & 47.71 / 77.53 & 38.91 / 68.32 \\
& OLDFS \cite{wang2024learning} & \blue{33.27} & 56.80 & \blue{12.69} & \blue{55.28} / - & \red{49.37} / - & \blue{41.10} / - \\
& PP-SSL (Ours) & \red{36.75} & \red{63.52} & \red{14.64} & \red{55.58/81.73} & \blue{49.30}/\red{76.90} & \red{41.38/68.83} \\
\bottomrule
\end{tabular}
}
\end{table*}

\subsection{Experimental Setup}
\textbf{Datasets.} We evaluate our proposed method on 7 public fine-grained image classification datasets, including CUB-200-2011 (200 bird species), Stanford Cars (196 car categories), FGVC-Aircraft (100 aircraft categories), NABirds (555 bird species), Flowers102 (102 flower species), Butterfly200 (200 butterfly species), and Stanford Dogs (120 dog breeds). Specifically, CUB-200-2011 \cite{wah2011caltech}: 11,788 images, 200 bird species, with 5,994 training and 5,794 testing images.
Stanford Cars \cite{krause20133d}: 16,185 images, 196 car categories, with 8,144 training and 8,041 testing images.
FGVC-Aircraft \cite{maji2013fine}: 10,000 images, 100 aircraft categories, with 6,667 training and 3,333 testing images.
NABirds \cite{van2015building}: 48,562 images, 555 bird species, with 23,929 training and 24,633 testing images.
Flowers102 \cite{nilsback2008automated}: 7,169 images, 102 flower species, with 1,020 training and 6,149 testing images.
Butterfly200 \cite{chen2018fine}: 25,279 images, 200 butterfly species, with 10,270 training and 15,009 testing images.
Stanford Dogs \cite{khosla2011novel}: 20,580 images, 120 dog breeds, with 12,000 training and 8,580 testing images.

\noindent\textbf{Implementation Details.} We employ the ResNet50 \cite{he2016deep} as the backbone of our network, initialized with ImageNet-trained weights. Following MoCo v2~\cite{chen2020improved}, the momentum factor of our MoCo contrastive module is set to 0.999. The projection head $g_{\theta}$ consists of two fully connected layers with ReLU activation and a linear layer with batch normalization (BN) \cite{ioffe2015batch}. We set the batch size to 128, and use the SGD optimizer with a learning rate of 0.03, momentum of 0.9, and weight decay of 0.0001. The CLIP image encoder (i.e., teacher model) and CLIP text encoder employ the viT-B/32 architecture. The retrieval phase is conducted over 100 epochs. During training, images in the FGVR dataset were resized to 224×224 pixels. In the testing phase, images are resized to 256 pixels and then center-cropped to obtain a final size of 224×224 pixels.

\subsection{Evaluation Protocols}
We evaluate our method in two settings: image retrieval and linear probing. First, we use image retrieval to assess the learned features by identifying images that match the query's category. This approach is crucial in unsupervised learning, as it relies on high-quality features without requiring extensive labeled data. Specifically, it effectively measures the features' ability in similarity retrieval, emphasizing its practicality as it requires no manual annotations or human intervention. We use rank-1 accuracy, rank-5 accuracy, and mean Average Precision (mAP) to provide a comprehensive assessment of feature quality. Secondly, linear probing is a common evaluation protocol for assessing the quality of features learned by SSL algorithms. In this setting, the SSL-trained feature extractor is fixed, and a linear classifier is trained on the extracted features. The classifier's performance reflects the quality and utility of the learned features for classification tasks.

\begin{table}[!h]
\footnotesize
\centering
\renewcommand{\arraystretch}{1.1}  
\caption{Comparison results between our method and other self-supervised learning methods on the CUB-200-2011, Stanford Cars, and FGVC Aircraft datasets. Retrieval accuracy (rank-1, in \%) and Top-1 accuracy (in \%) based on linear classification with frozen feature extractor representations are reported.}
\label{tab:2}
\setlength{\tabcolsep}{0.6mm}{
\begin{tabular}{l|ccc|ccc}
\toprule
\multirow{2}{*}{Method}  & \multicolumn{3}{c|}{Image Retrieval} & \multicolumn{3}{c}{Classification} \\
\cmidrule(lr){2-4} \cmidrule(lr){5-7}
& CUB & Cars & Aircraft & CUB & Cars & Aircraft \\
\midrule

supervised & - & - & - & 77.46 & 88.60 & 85.93\\
\midrule
Dino~\cite{chen2020simple} & - & - & - & 16.74 & 14.33 & 12.07\\
Simsiam~\cite{chen2021exploring} & 16.24 & 12.45 & 18.49 & 46.75 & 45.72 & 38.52 \\
MoCo V2~\cite{chen2020improved} & 39.72 &30.51 &30.02 &63.98 &62.02 &51.13 \\
DiLo~\cite{zhao2021distilling} & - & - & - & 62.97 & - & -\\
CVSA~\cite{wu2021exploring} & - & - & - & 63.02 & - & -\\
LEWEL~\cite{huang2022learning} & 39.91 &32.36 &31.09 &64.59 &62.91 &51.90 \\
ContrastiveCrop~\cite{peng2022crafting} & 39.84 & 32.71 &30.37 &64.23 &63.29 &52.04\\
SAM-SSL-Bilinear~\cite{shu2022improving} &40.08 &33.19 &30.52&64.94 &62.85 &52.83 \\
LCR~\cite{shu2023learning} & \blue{44.41} & \blue{36.46} &32.97 &65.19&65.54 &54.01 \\
OLDFS~\cite{wang2024learning} &42.06 &35.81 & \blue{33.27} & \blue{66.17} & \blue{65.60} & \blue{55.28} \\
\midrule
Ours & \red{53.19} & \red{41.25} & \red{36.75} & \red{69.26} & \red{67.73} & \red{55.85} \\
\bottomrule
\end{tabular}
}
\end{table}

\subsection{Experimental Results}

\textbf{Effectiveness of the Proposed Method.} To evaluate the performance improvements of our method, we first compared it with two recent advanced methods, i.e., LCR and OLDFS, for retrieval and classification tasks. As illustrated in Tab.~\ref{tab:1}, our method significantly outperforms other methods. On the CUB-200-2011 dataset, our method achieved the best performance in various label proportions for classification tasks. Besides, on the Stanford Cars and FGVC Aircraft datasets, our method achieved the highest performance in terms of rank-1 and rank-5 for retrieval tasks. Specifically, our method improved the rank-1 accuracy by 8.78\%, 4.79\%, and 3.48\% over two advanced methods on the three datasets, with a particularly notable improvement of 8.78\% on CUB-200-2011. Our method’s superior performance in the retrieval task is attributed to the integration of AIS and IADM, which effectively mitigate the interference of irrelevant features and harness discriminative cues from the original image, thus driving improvements in fine-grained retrieval tasks. In terms of classification metrics, our method also demonstrates performance gains. However, in certain datasets and label proportion settings, the OLDFS method does not show a significant gap compared to our method. This may be due to OLDFS’s ability to learn task-irrelevant features, which could contribute to enhancing performance in downstream visual recognition tasks \cite{chen2021exploring, li2022understanding}.

\begin{table*}[!t]
\centering
\renewcommand{\arraystretch}{1.15}  
\caption{All models utilize ResNet-50 as the network backbone, with the architecture initialized using ImageNet-pretrained weights. We conduct comparisons with other self-supervised methods across four additional FGVR datasets. For both image retrieval and image classification tasks, the batch size is set to 128. Retrieval accuracy is reported as rank-1/rank-5 (in \%), and classification accuracy is presented as top-1/top-5 (in \%).}
\setlength{\tabcolsep}{4mm} 
\label{tab:3}
\resizebox{\textwidth}{!}{
\begin{tabular}{c|cc|cc|cc|cc}
\toprule
{\multirow{2}{*}{Method}} & \multicolumn{2}{c|}{Stanford Dog}   & \multicolumn{2}{c|}{Flowers-102}  & \multicolumn{2}{c|}{Butterfly-200}   & \multicolumn{2}{c}{Nabird}  \\ \cmidrule(lr){2-3} \cmidrule(lr){4-5} \cmidrule(lr){6-7} \cmidrule(lr){8-9}
  & Retrieval   & Classification      & Retrieval   & Classification  & Retrieval & Classification & Retrieval & Classification  \\
\midrule
SimSiam \cite{chen2021exploring}  & 27.56/41.45  & 58.64/74.18 & 34.13/58.36  & 62.14/77.46   & 24.97/39.45 & 57.59/76.82  & 13.53/21.63   & 41.91/66.42    \\
\midrule
Dino \cite{zhao2021distilling}  & -   & 32.48/42.63  & -  & 41.56/49.67  & -  & 31.81/40.88   & -  & 14.74/20.23   \\
\midrule
MoCo V2~\cite{chen2020improved}  & 69.57/87.81  & 82.57/93.14   & 88.46/94.78      & 88.12/92.93   & 70.58/87.02  & 77.64/82.51   & 33.67/57.45  & 54.26/77.84  \\
\midrule
LCR \cite{shu2023learning}  & 74.48/91.33 & 84.42/98.19  & 92.91/97.53 & 90.45/97.95 & 71.62/89.00 & 80.23/96.55 & 36.52/61.06 & 55.24/81.24 \\
\midrule
Ours & \red{75.58/91.74} & \red{85.09/98.86} & \red{93.48/98.08} & \red{91.19/98.03} & \red{73.07/90.45} & \red{80.94/96.85} & \red{41.62/66.44} & \red{57.80/82.97} \\
\bottomrule
\end{tabular}
}
\end{table*}

\noindent \textbf{Comparison with Other SSL Methods.} Furthermore, we compared our method with other self-supervised learning approaches to evaluate its performance in fine-grained recognition tasks. We report the rank-1 accuracy for image retrieval and top-1 accuracy for classification, with all experiments conducted using a batch size of 128, as shown in Tab.~\ref{tab:2}. Our method consistently achieves the highest rank-1 and top-1 accuracies on the CUB-200-2011, Stanford Cars, and FGVC Aircraft datasets. It demonstrates sustained competitiveness in both retrieval and classification tasks compared to other SSL methods. Compared to the latest self-supervised approaches, our method continues to exhibit outstanding performance. 

\begin{figure*}[!t]
    \centering
    \includegraphics[width=1\linewidth]{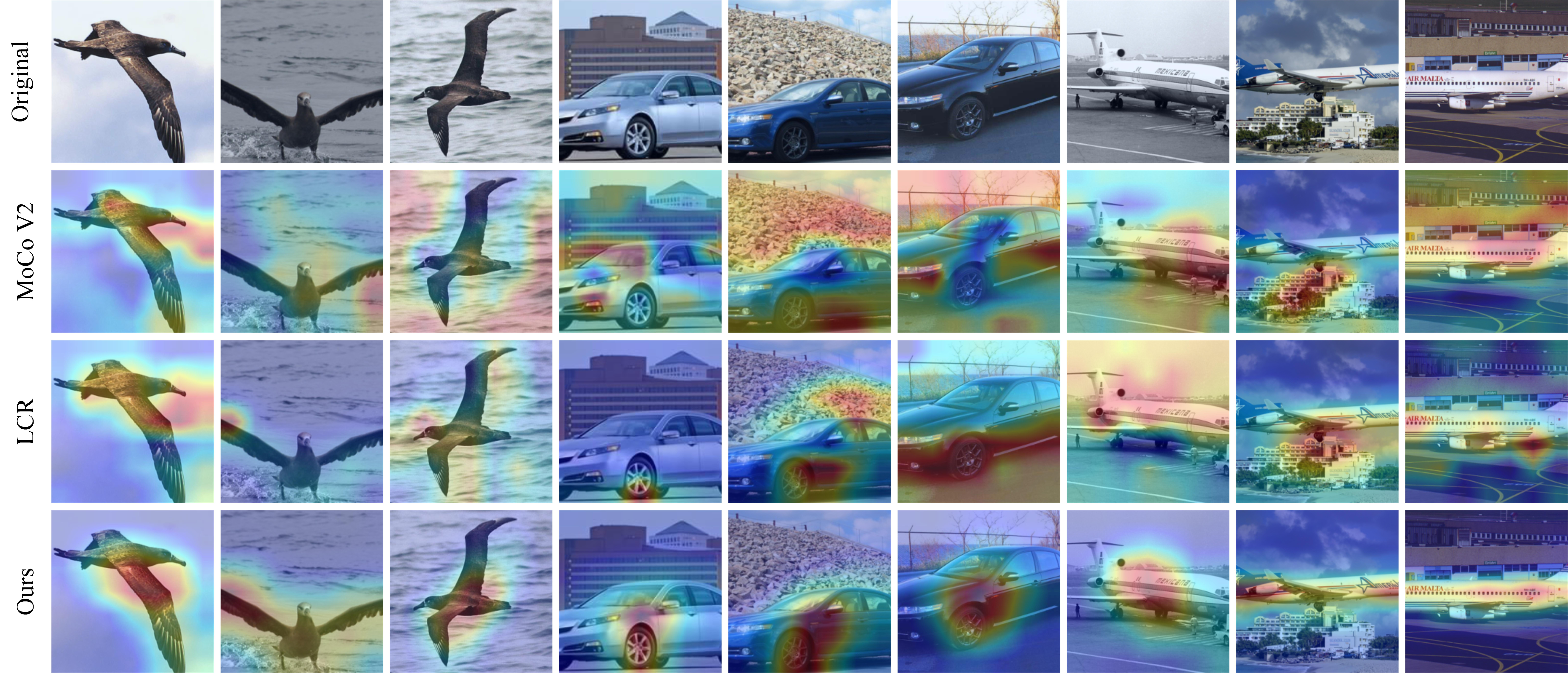}
    \caption{Attention map visualizations on the CUB-200-2011, Stanford Cars, and FGVC Aircraft datasets comparing our method with others. Our method effectively reduces interference from irrelevant features and identifies key parts of the target object.}
    \label{attn_Visualisation}
\end{figure*}

We further conducted experiments on four public FGVR datasets. As shown in Tab.~\ref{tab:3}, our method achieves the best performance across these datasets as well. Fig.~\ref{attn_Visualisation} presents visualizations of the attention regions for our method and others. By visualizing the regions the model attends to, our method shows an enhanced ability to focus on more discriminative cues while diminishing the impact of irrelevant features, leading to superior performance.

\begin{table}[!h]
    \centering
    \footnotesize
    \renewcommand{\arraystretch}{1.15} 
    \caption{We conducted ablation experiments on the CUB-200-2011 dataset and reported the rank-1, rank-5, and mAP (in \%) performance for the retrieval task.}
\label{tab:4}
\setlength{\tabcolsep}{1.8mm}{
\begin{tabular}{c|ccccccc}
\toprule
layer0 & $\checkmark$ &  &  &  & & $\checkmark$ & $\checkmark$ \\
layer1 & & $\checkmark$ & &  & &  &  \\
layer2 &  &  & $\checkmark$ &  &  &  & \\
layer3 & & &  & $\checkmark$ &  &  &    \\
layer4 & & &  & & $\checkmark$ & $\checkmark$ & \\
$AIS$  &   & $\checkmark$ & $\checkmark$ & $\checkmark$ & $\checkmark$ & $\checkmark$ & $\checkmark$ \\
\midrule
rank-1 &  46.91  & 52.21 & 52.17 & 52.81 & 47.13   & 50.36  & \red{53.19}  \\ \cline{1-8}
rank-5 & 71.11  & 77.44 & 77.56 & 78.25   & 74.06  & 76.91  & \red{78.32} \\ \cline{1-8}
map   &  20.43 & 26.57 & 26.45 & 26.50 & 21.91 & 24.99  & \red{26.31}   \\
\bottomrule
\end{tabular}}

\end{table}

\begin{table}[!h]
\footnotesize
\centering
\renewcommand{\arraystretch}{1.0}  
\caption{We conducted ablation experiments on text description using the CUB-200-2011 dataset and reported the rank-1, rank-5, and mAP (in \%) performance for the retrieval task.} 
\label{tab:5}
\setlength{\tabcolsep}{0.8mm}{
\footnotesize
\begin{tabular}{c|c|ccc}
\toprule
{\multirow{2}{*}{Method}} & {\multirow{2}{*}{Description}} & \multicolumn{3}{c}{Image Retrieval}  \\
\cmidrule(lr){3-5}
&  & rank-1 & rank-5 & mAP  \\
\midrule

Coarse-Grained Text & ``a bird'' & 51.73 & 77.89 & 26.34 \\

\midrule
{\multirow{3}{*}{Fine-Grained Text}} & ``an animal characterized &  &  &  \\
& by feathers, wings, and & \red{53.19} & \red{78.32} & \red{26.31} \\
& the ability to fly or perch'' &  &  &  \\

\bottomrule
\end{tabular}}

\end{table}

\subsection{Ablation Study of PP-SSL Architecture}
The ablation experiments are conducted on the CUB-200-2011 dataset, with results for other datasets provided in the supplementary material.

\begin{figure*}[!t]
    \centering
    \includegraphics[width=1.0\linewidth]{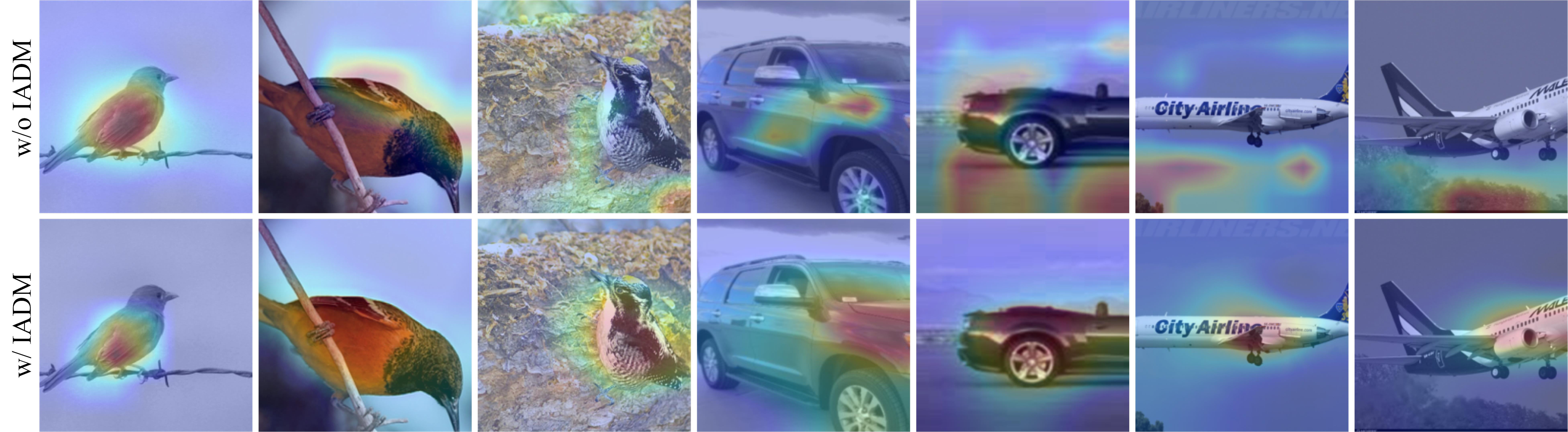}   
    \caption{The effectiveness of the proposed IADM is shown via GradCAM visualization, highlighting finer discriminative features identified in the image.}
    \label{IADM_Visualisation}
\end{figure*}
\noindent \textbf{Anti-Interference Strategy (AIS).}
As shown in the first and last columns of Tab.~\ref{tab:4}, applying AIS on top of layer$_0$ (i.e., IADM, guided by the original image information) significantly improves the retrieval rank-1 and rank-5 accuracy on the CUB-200-2011 fine-grained dataset, with increases of 6.28\% and 7.21\%, respectively. This substantial improvement highlights the effectiveness of the proposed AIS in mitigating interference from irrelevant features. Additionally, we conducted ablation studies on two other datasets, exploring various combination strategies, with the results provided in the appendix. 

\noindent \textbf{Ablation of the Number of AIS Fine-Grained Descriptions.}
Tab.~\ref{tab:5} shows the performance differences between using coarse class text and fine-grained description text. The fine-grained descriptions are more effective in mitigating interference from irrelevant features.

\noindent \textbf{Image-Aided Distinction Module (IADM).}
The GradCAM visualizations with and without IADM are shown in Fig.~\ref{IADM_Visualisation}. It can be observed that interference from irrelevant regions is significantly reduced, enhancing the model's ability to capture fine-grained discriminative features within key areas and focus on more detailed distinguishing patterns, thereby demonstrating the effectiveness of the proposed IADM. Furthermore, as shown in Tab.~\ref{tab:4}, the GradCAM obtained by computing gradients with respect to the original image (i.e., IADM) achieves the best performance compared to using other layers for guidance (from columns 2 to the last in Tab.~\ref{tab:4}). Notably, the combination of deep-layer features and original image guidance is less effective than using original image guidance alone. The combination of AIS and IADM achieves the best performance.

\subsection{Further Analysis}

\begin{table}[!h]
    \centering
    \footnotesize
    \renewcommand{\arraystretch}{0.95} 
    \caption{Hyperparameter anaylsis in terms of rank-1, rank-5, and mAP (all in \%) on the CUB-200-2011 dataset.}
\label{tab:6}
\setlength{\tabcolsep}{4.3mm}{
    \begin{tabular}{c|c|c|c}
\toprule
        Weights & rank-1 & rank-5 & mAP \\ 
        \midrule
        $\alpha=1.2,\beta=0.009$ & 52.01 & 76.98 & 25.16 \\ 
        $\alpha=1,2,\beta=0.01$ & \red{53.19} & \red{78.32} & \red{26.31} \\ 
        $\alpha=1.2,\beta=0.2$ & 51.34 & 76.17 & 24.93 \\ 
        $\alpha=1.0,\beta=0.01$ & 50.62 & 74.23 & 23.26 \\ 
        $\alpha=1.4,\beta=0.01$ & 51.87 & 74.92 & 24.20 \\ 
        \bottomrule
    \end{tabular}}
\end{table}

\noindent \textbf{Analysis of Hyperparameters.} In this section, we conduct sensitivity analysis of two hyperparameters, i.e., $\alpha$ and $\beta$ used in Eq.~\ref{total+loss}, on the CUB-200-2011 dataset, which determine the strength of the weights of $\mathcal{L}_{AIS}$ and $\mathcal{L}_{IADM}$, respectively. The analysis results are demonstrated in Tab.~\ref{tab:6}. It is observed that $\beta=1.2$ and $\gamma=0.01$ achieve the best performance. Therefore, we adopt it for our experiments.

\begin{figure}[!h]
    \centering
    \includegraphics[width=1.0\linewidth]{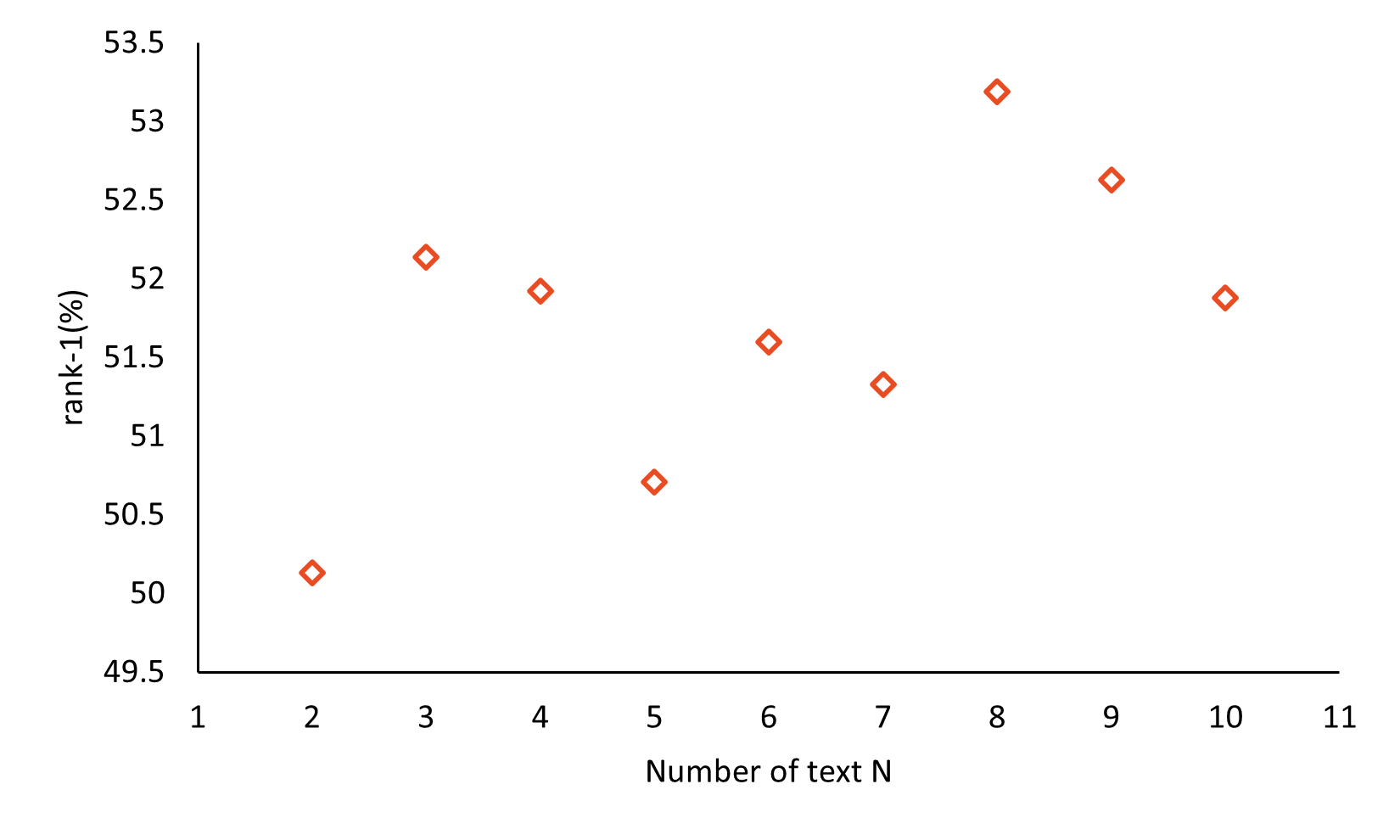}   
    \caption{Analysis of the text number $N$ in terms of Rank-1 metric (in \%) on the CUB-200-2011 Dataset.}
    \label{number of n}
\end{figure}

\noindent \textbf{Analysis of the Preset Text Library $N$.} Here, we further analyze the effect of the number of text prompts ($N$) in the preset text library. As shown in Fig.~\ref{number of n}, storing 8 text prompts achieves the highest rank-1 accuracy. Therefore, we set $N=8$ in this paper by default. For the text prompt configuration within the library, we employed a more refined prompt, like “an animal characterized by feathers, wings, and the ability to fly or perch.” Future experiments will explore alternative designs for these prompts.
\section{Conclusion}
This paper presents PP-SSL, a novel self-supervised framework for fine-grained visual recognition, addressing the issues of irrelevant feature interference and mitigating granularity bias. Specifically, the proposed anti-interference strategy enables the model to acquire semantic understanding of categories, allowing it to focus on key regions of the target while reducing the impact of irrelevant feature interference in fine-grained visual recognition tasks. Additionally, the proposed image-aided distinction module extracts crucial fine-grained cues, enhancing the model's ability to distinguish subtle differences. Extensive experiments on 7 benchmarks show that our PP-SSL outperforms recent state-of-the-art methods in both classification and retrieval tasks.

{
    \small
    \bibliographystyle{ieeenat_fullname}
    \bibliography{main}
}


\end{document}